\documentclass[letterpaper, 10 pt, conference]{ieeeconf}  

\IEEEoverridecommandlockouts                              
\usepackage{graphicx}
\usepackage{epsfig} 
\usepackage{mathptmx} 
\usepackage{times} 
\usepackage{amsmath} 
\usepackage{amssymb}  
\usepackage{cite}
\usepackage{tikz}

\usepackage[normalem]{ulem}
\useunder{\uline}{\ul}{}

\title{\LARGE \bf
Predicting Yelp Star Reviews Based on Network Structure with Deep Learning}

\author{Luis A. Perez$^{1}$
\thanks{$^{1}$L. Perez is an MS Candidate in the School of Engineering at Stanford University,
        450 Serra Mall, Stanford, CA 94305, USA
        {\tt\small luis0 at stanford.edu}}%
}

\begin{document}

\maketitle
\thispagestyle{empty}
\pagestyle{empty}

\begin{abstract}

In this paper, we tackle the real-world problem of predicting Yelp star-review rating based on business features (such as images, descriptions), user features (average previous ratings), and, of particular interest, network properties (which businesses has a user rated before). We compare multiple models on different sets of features -- from simple linear regression on network features only to deep learning models on network and item features.

In recent years, breakthroughs in deep learning have led to increased accuracy in common supervised learning tasks, such as image classification, captioning, and language understanding. However, the idea of combining deep learning with network feature and structure appears to be novel. While the problem of predicting future interactions in a network has been studied at length, these approaches have often ignored either node-specific data or global structure \cite{PintrestProject}.

We demonstrate that taking a mixed approach combining both node-level features and network information can effectively be used to predict Yelp-review star ratings. We evaluate on the Yelp dataset by splitting our data along the time dimension (as would naturally occur in the real-world) and comparing our model against others which do no take advantage of the network structure and/or deep learning.

\end{abstract}

\section{Introduction}
The problem of predicting network structure can be both of great practical importance as well as a case-study in understanding the usefulness of deep learning in network settings. An accurate model can be used to suggest friend recommendations, product recommendations, and even predict individual user actions. A system which solves this problem is generally referred to in the literature as a recommender system, and such systems are quite common at large Internet companies such as Amazon \cite{Linden:2003:ARI:642462.642471}, Netflix \cite{Zhou:2008:LPC:1424237.1424269}, and Google.

The main approaches typically taken fall into two categories - \textit{content based} and \textit{collaborative filtering} approaches. The first makes use of text, meta-data, and other features in order to identify potentially related items, while the latter leans more towards making use of aggregated behavior and of a large number of training samples (ie, users and businesses). Collaborative filtering approaches have proven useful in recommender systems in industry, and are typically the preferred method due to how expensive it typically is (in both computational resources and engineering effort) to extract useful features from large amounts of meta-data. However, with advances in deep learning (extracting features from videos and text that are useful for many tasks), it seems feasible that revisiting content-based approaches with additional network-level data will prove fruitful.

In this paper, we seek to explore a novel method combining both deep learning feature extraction (a \textit{content-based} approach) with network prediction models (a quasi-\textit{collaborative filtering} approach). We focus on a real-world, practical network - the Yelp Review Network. The network consists of 4.7M review (edges), 156K businesses, 200K pictures, covering over 12 metropolitan areas in the united state.

Specifically, we seek to model the problem of predicting a user's star rating of a previously unrated business by using features about the business, the user, as well as existing interactions between the user and other businesses.

From a general view point, we hypothesize that the final star rating given by a users is a mixture of all of the above interactions. In particular, we would expect that rating at time $t$ between user $i$ and business $j$ could be modeled as:
$$
r_t = f(i_t, j_t, \delta_{i,j,t}) + \mathcal{N}(0,\epsilon_{i,j,t})
$$

Here, we have $i_t$ is the overall user-based bias at time $t$. For example, some users simply tend to give higher or lower ratings based on previous experience -- one could argue this is inherent to the user directly. We also have $j_t$, the overall business bias at time $t$. For example, some business are objectively better across the board, by having better food, websites, or being at better locations. Finally, the term $\delta_{i,j,t}$ which is an interaction term reflecting the interaction between this user and the business as time $t$. One might imagine for example that a user who really enjoys Mexican food will tend to give those restaurants a higher rating.

In the end, these three terms should be combined in some way (with normalization, etc.) to arrive at a final rating. As such, we essentially have four models which can be combined to give better predictive power:

\begin{itemize}
\item a user model, trained only on user properties
\item a business model, trained on business properties
\item interaction model trained on a mixture of both properties with additional features known only to the network (such as previous business interactions, etc).
\end{itemize}

\section{Related Work}
In general, there are three areas of interest in the literature. We have (1) work which focuses and provides techniques for predicting results based on network structures, (2) work which has applied some ML techniques to the features extracted from networks (and sometimes elements themselves), and (3) work which throws away a lot of the network structure and focuses exclusively on using the data to make predictions. All of these are supervised learning methods which varying degrees of complexity. We provide a brief overview of them, followed by a section discussing the mathematical underpinnings of the models.

\subsection{Graph-Based Approaches}

Liben-Nowell and Kleinberg \cite{TheLinkPredictionProblemForSocialNetworks} formalize the \textit{link prediction problem} and develop a proximity-based approach to predict the formation of links in a large co-authorship network. The model focuses on the network topology alone, ignoring any additional meta-data associated with each node since its basic hypothesis is that the known network connections offer sufficient insight to accurately predict network growth over time. They formally tackle the problem of given a social graph $G = (V,E)$ where each edge represents an interaction between $u,v$ and a particular timestamps $t$, can we use a subset of the graph across time (ie, with edges only in the interval $[t,t']$ to predict a future subset of the graph $G'$). The methods presented ignore the creation of new nodes, focusing only on edge prediction.

Multiple predictors $p$ are presented, each focusing on only network structure. For example, some intuitive predictors (there are many others studied, though not necessarily as intuitive) for the edge creation between $x$ and $y$:

\begin{enumerate}
\item graph distance -- (negated) length of the shortest path between $x$ and $y$
\item preferential attachments -- $|\Gamma(x)| \cdot |\Gamma(y)|$ where $\Gamma: V \to 2^V$ is a map from nodes to neighbors of nodes.
\end{enumerate}

Each of the above predictors $p$ can output a ranked list of most likely edges. The paper evaluates effectiveness by comparing calculating the percentage of edges which are correctly predicted to exists in the test data. The baseline for the paper appears to be a random predictor based on the training graph and the graph distance predictor. The predictors are evaluated over five difference co-authoring networks. =

The predictors can be classified into essentially three categories:

\begin{itemize}
\item Predictors based on local network structure
\item Predictors based on global network structure
\item Meta predictors based on a mixture of the above two 
\end{itemize}

All predictors performed above the random baseline, on average. The hitting time predictors performed below the graph distance baseline, with a much narrower positive gap for the remaining predictors. Most predictors performed on-par with just a common neighbors predictors.

\subsection{Introducing ML}

Further work by Leskovec et al. \cite{Leskovec:2010:PPN:1772690.1772756} seeks to introduce the nuance of both ``positive'' and ``negative'' relationships to the link prediction problem, addressing limitations of previous work. In concrete, it seeks to predict the sign of each edge in a graph based on the local structure of the surrounding edges. Such predictions can be helpful in determining future interactions between users, as well as determining polarization of groups and communities. 

Leskovec et al. introduce the ``edge sign prediction problem'' and study it in three social networks where explicit trust/distrust is recorded as part of the graph structure, work which is later expanded by Chiang et al. \cite{Chiang:2011:ELC:2063576.2063742}. The explicit sign of the edges is given by a vote for or a vote against, for example, in the Wikipedia election network. They find that their prediction performance degrades only slightly across these three networks, even when the model is trained on one network and evaluated against another.

They also introduces social-psychological theories of balance and status and demonstrates that these seems to agree, in some predictions, with the models explored.

Furthermore, they introduces the novel idea of using a machine learning approach built on top of the network features to improve the performance of the model. Rather than rely directly on any one network features, it instead extracts these features from the network and uses them in a machine learning model, achieving great performance. The features selected are, roughly speaking:

\begin{itemize}
\item Degree features for pair $(u,v)$ - there are seven such features, which are (1) the number of incoming positive edges to $v$, (2) the number of incoming negative edges to $v$, (3) the number of outgoing positive edges from $u$, (4) the number of outgoing negative edges from $u$, (5) the total number of common neighbors between $u$ and $v$, (6) the out-degree of $u$ and the (7) in-degree of $v$.
\item Triad features - We consider 16 distinct triads produced by $u,v,w$ and count how many of each type of triad.
\end{itemize}

The above features are fed into a logistic regression model and are used to relatively successfully predict the sign of unknown edges.

Overall, while previous network predictions problems have attempted to make use of machine learning, most still rely on relatively simple models and have not yet made the jump to deeper architectures.

\subsection{Content-Based Deep Learning}
Hasan et. al in \cite{Hasan06linkprediction} introduce the very important idea of using features of the node to assist in link prediction. The paper also significantly expands on the set of possible models to use for ML, demonstrating that for their data, SVMs work the best when it comes to predicting the edge. They formulate their problem as a supervised machine learning problem. Formally, we take two snapshots of a network at different times $t$ and $t'$ where $t' > t$. The training set of generated by choosing pairs of nodes $(u,v)$ which are not connected by an edge in $G_t$, and labeling as positive if they are connected in $G_{t'}$ and negative if they are not connected in $G_{t'}$. The task then becomes a classification problem to predict whether the edges $(u,v)$ is positive or negative. 

In particular,they make use of the following features:

\begin{itemize}
\item Proximity features - computed from the similarity between nodes.
\item Aggregated features - how "prolific" a scientists is, or other features that belong to each node.
\item Network topology features - (1) shortest distance among pairs of nodes, (2) number of common neighbors, (3) Jaccard's coefficient, etc.
\end{itemize}

The authors rigorously describes the sets of features it found the most predictive, and takes into account node-level information extractable from the network as well as some amount of ``meta''-level information (for example, how similar two nodes are to each other). The results demonstrate great success (with accuracies up to 90\% compared to a baseline of 50\% or so). Overall, The authors presents a novel approach of using machine learning to assist in the link prediction problem by rephrasing the problem as a supervised learning task.

\section{Methodology and Data}
In this section, we describe the architecture of our feature extraction networks as well as lay the ground work for our predictive models. We define our loss function and presents some additional details used for training, such as learning rate and other hyper-parameters.

We convert the original data from JSON format to CSV. The data set contains 156,639 businesses (with 101 distinct attributes), 196,278 photos (associated with businesses), 1,028,802 tips (these are between users and businesses), 135,148 check-ins (again, associated with each business), and 1,183,362 users.

\subsection{Dataset}
Our dataset is the set released for the Yelp Data Set Challenge Round 10 \cite{YelpDataSet} in 2017. The entirety of the dataset consists of the following entities:
\begin{itemize}
\item \textbf{Businesses}: Consists of exactly 156,639 businesses. It contains data about businesses on Yelp including geographical location, attributes, and categories.
\item \textbf{Reviews}: 4,736,897 reviews. It contains full review text (for NLP processing) as well as the user id that wrote the review and the business id the review is written for. It also contains the number of stars given, as well as the number of useful, funny, and cool up-votes (finally, it also contains the date).
\item \textbf{Users}: 1,183,362 Yelp users. It includes the user's friend mapping and all the meta-data associated with the user. Just this single dataset consists contains 538,440,966 edges.
\item \textbf{Tips}: 1,028,802 tips. Tips are associated with each business and are written by users. Tips are similar to reviews, but without rating and usually much shorter.
\item \textbf{Photos:} 196,278 Photos, each associated with businesses. The photos are also associated with captions.
\item \textbf{Check-ins:} 135,148 check-ins on a business (this a business only attribute).
\end{itemize}

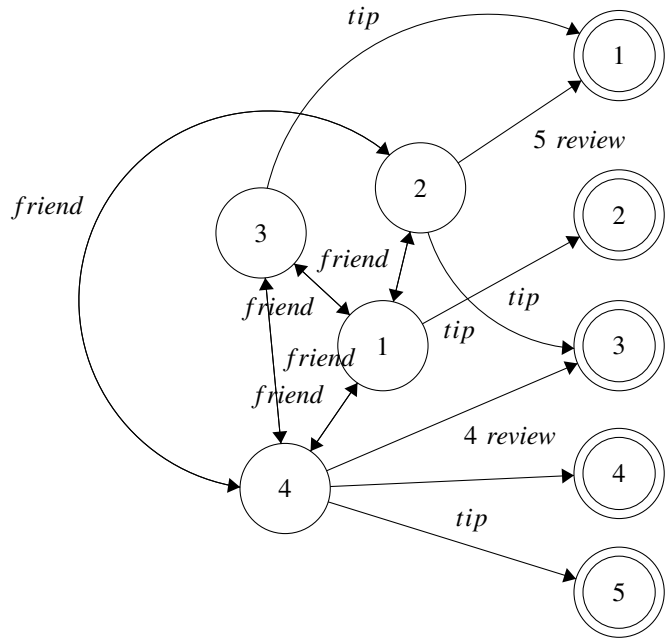
\begin{figure}[h!]
\centering
\begin{tikzpicture}[scale=0.2]
\tikzstyle{every node}+=[inner sep=0pt]
\draw [black] (28.4,-27.4) circle (3);
\draw (28.4,-27.4) node {$1$};
\draw [black] (30.9,-16.9) circle (3);
\draw (30.9,-16.9) node {$2$};
\draw [black] (20.3,-19.9) circle (3);
\draw (20.3,-19.9) node {$3$};
\draw [black] (21.9,-36.9) circle (3);
\draw (21.9,-36.9) node {$4$};
\draw [black] (44.1,-8.1) circle (3);
\draw (44.1,-8.1) node {$1$};
\draw [black] (44.1,-8.1) circle (2.4);
\draw [black] (44.1,-18.7) circle (3);
\draw (44.1,-18.7) node {$2$};
\draw [black] (44.1,-18.7) circle (2.4);
\draw [black] (44.1,-27.4) circle (3);
\draw (44.1,-27.4) node {$3$};
\draw [black] (44.1,-27.4) circle (2.4);
\draw [black] (44.1,-35.9) circle (3);
\draw (44.1,-35.9) node {$4$};
\draw [black] (44.1,-35.9) circle (2.4);
\draw [black] (44.1,-43.8) circle (3);
\draw (44.1,-43.8) node {$5$};
\draw [black] (44.1,-43.8) circle (2.4);
\draw [black] (33.4,-15.24) -- (41.6,-9.76);
\fill [black] (41.6,-9.76) -- (40.66,-9.79) -- (41.22,-10.62);
\draw (41.47,-13) node [below] {$5\mbox{ }review$};
\draw [black] (41.115,-27.585) arc (-94.24841:-162.7529:11.037);
\fill [black] (41.11,-27.58) -- (40.35,-27.03) -- (40.28,-28.02);
\draw (33.5,-25.71) node [below] {$tip$};
\draw [black] (30.21,-19.82) -- (29.09,-24.48);
\fill [black] (29.09,-24.48) -- (29.77,-23.82) -- (28.79,-23.59);
\draw (28.89,-21.73) node [left] {$friend$};
\draw [black] (29.09,-24.48) -- (30.21,-19.82);
\fill [black] (30.21,-19.82) -- (29.53,-20.48) -- (30.51,-20.71);
\draw [black] (22.5,-21.94) -- (26.2,-25.36);
\fill [black] (26.2,-25.36) -- (25.95,-24.45) -- (25.27,-25.19);
\draw (21.45,-24.14) node [below] {$friend$};
\draw [black] (26.2,-25.36) -- (22.5,-21.94);
\fill [black] (22.5,-21.94) -- (22.75,-22.85) -- (23.43,-22.11);
\draw [black] (26.71,-29.88) -- (23.59,-34.42);
\fill [black] (23.59,-34.42) -- (24.46,-34.05) -- (23.63,-33.48);
\draw (24.55,-30.8) node [left] {$friend$};
\draw [black] (23.59,-34.42) -- (26.71,-29.88);
\fill [black] (26.71,-29.88) -- (25.84,-30.25) -- (26.67,-30.82);
\draw [black] (21.62,-33.91) -- (20.58,-22.89);
\fill [black] (20.58,-22.89) -- (20.16,-23.73) -- (21.15,-23.64);
\draw (21.73,-28.32) node [right] {$friend$};
\draw [black] (20.58,-22.89) -- (21.62,-33.91);
\fill [black] (21.62,-33.91) -- (22.04,-33.07) -- (21.05,-33.16);
\draw [black] (18.908,-36.812) arc (-98.51807:-309.93742:12.583);
\fill [black] (18.91,-36.81) -- (18.19,-36.2) -- (18.04,-37.19);
\draw (8.57,-18.2) node [left] {$friend$};
\draw [black] (18.908,-36.811) arc (-98.5328:-309.92269:12.582);
\fill [black] (28.85,-14.72) -- (28.56,-13.82) -- (27.92,-14.59);
\draw [black] (31.02,-25.95) -- (41.48,-20.15);
\fill [black] (41.48,-20.15) -- (40.53,-20.1) -- (41.02,-20.98);
\draw (37.8,-23.55) node [below] {$tip$};
\draw [black] (20.708,-16.933) arc (166.48658:66.25769:15.116);
\fill [black] (41.49,-6.63) -- (40.96,-5.85) -- (40.56,-6.76);
\draw (27.15,-6.41) node [above] {$tip$};
\draw [black] (24.9,-36.77) -- (41.1,-36.03);
\fill [black] (41.1,-36.03) -- (40.28,-35.57) -- (40.33,-36.57);
\draw [black] (24.76,-37.79) -- (41.24,-42.91);
\fill [black] (41.24,-42.91) -- (40.62,-42.19) -- (40.32,-43.15);
\draw (34.33,-39.78) node [above] {$tip$};
\draw [black] (24.66,-35.72) -- (41.34,-28.58);
\fill [black] (41.34,-28.58) -- (40.41,-28.44) -- (40.8,-29.35);
\draw (36.85,-32.7) node [below] {$4\mbox{ }review$};
\end{tikzpicture}
\caption{Proposed Complex Graph Models Based on Users, Reviews, Businesses, User-User Interactions, and Tips}
\label{fig:graph_complex_structure}
\end{figure}

As we can see from above, the dataset is relatively rich, with many possible graph structures to study on top of it. In general, given that we are trying to predict review ratings, we focus on the following bipartite graph with users and businesses:

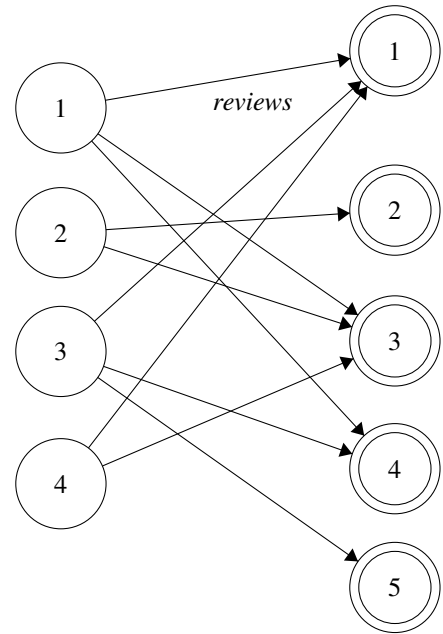
\begin{figure}[h!]
\centering
\begin{tikzpicture}[scale=0.2]
\tikzstyle{every node}+=[inner sep=0pt]
\draw [black] (21.9,-11.9) circle (3);
\draw (21.9,-11.9) node {$1$};
\draw [black] (21.9,-20.2) circle (3);
\draw (21.9,-20.2) node {$2$};
\draw [black] (21.9,-28.1) circle (3);
\draw (21.9,-28.1) node {$3$};
\draw [black] (21.9,-36.9) circle (3);
\draw (21.9,-36.9) node {$4$};
\draw [black] (44.1,-8.1) circle (3);
\draw (44.1,-8.1) node {$1$};
\draw [black] (44.1,-8.1) circle (2.4);
\draw [black] (44.1,-18.7) circle (3);
\draw (44.1,-18.7) node {$2$};
\draw [black] (44.1,-18.7) circle (2.4);
\draw [black] (44.1,-27.4) circle (3);
\draw (44.1,-27.4) node {$3$};
\draw [black] (44.1,-27.4) circle (2.4);
\draw [black] (44.1,-35.9) circle (3);
\draw (44.1,-35.9) node {$4$};
\draw [black] (44.1,-35.9) circle (2.4);
\draw [black] (44.1,-43.8) circle (3);
\draw (44.1,-43.8) node {$5$};
\draw [black] (44.1,-43.8) circle (2.4);
\draw [black] (24.86,-11.39) -- (41.14,-8.61);
\fill [black] (41.14,-8.61) -- (40.27,-8.25) -- (40.44,-9.23);
\draw (34.64,-10.84) node [below] {$reviews$};
\draw [black] (24.36,-13.62) -- (41.64,-25.68);
\fill [black] (41.64,-25.68) -- (41.27,-24.81) -- (40.7,-25.63);
\draw [black] (23.94,-14.1) -- (42.06,-33.7);
\fill [black] (42.06,-33.7) -- (41.89,-32.77) -- (41.15,-33.45);
\draw [black] (24.89,-20) -- (41.11,-18.9);
\fill [black] (41.11,-18.9) -- (40.27,-18.46) -- (40.34,-19.46);
\draw [black] (24.75,-21.13) -- (41.25,-26.47);
\fill [black] (41.25,-26.47) -- (40.64,-25.75) -- (40.33,-26.7);
\draw [black] (24.66,-35.72) -- (41.34,-28.58);
\fill [black] (41.34,-28.58) -- (40.41,-28.44) -- (40.8,-29.35);
\draw [black] (24.35,-29.83) -- (41.65,-42.07);
\fill [black] (41.65,-42.07) -- (41.29,-41.2) -- (40.71,-42.01);
\draw [black] (24.73,-29.09) -- (41.27,-34.91);
\fill [black] (41.27,-34.91) -- (40.68,-34.17) -- (40.35,-35.11);
\draw [black] (23.73,-34.52) -- (42.27,-10.48);
\fill [black] (42.27,-10.48) -- (41.38,-10.8) -- (42.18,-11.41);
\draw [black] (24.13,-26.09) -- (41.87,-10.11);
\fill [black] (41.87,-10.11) -- (40.94,-10.27) -- (41.61,-11.01);
\end{tikzpicture}
\caption{Simplified graph model of user reviews of businesses. The graph is bipartite, with users and businesses connected by directed "review" edges.}
\label{fig:graph_structure}
\end{figure}

and further propose making use of the friend-friend explicit graph (it is possible that it might be meaningful to see if we can find any relationship between friend reviews and user reviews) and the tip edges (without ratings, but possible meaningful information about a business). With this additional information, the structure of the graph itself becomes increasingly complex, as shown in Diagram \ref{fig:graph_complex_structure}.

\subsection{Predictive Models}
The rich meta-data about the network makes it quite interested to analyze, and opens up a lot of venues for possible improvements in terms of link prediction. We have multiple networks available for explorations, including \textit{user-user} network (based on friendships, comments, etc.), \textit{user-business} network, based on reviews given by a specific business to a user.

Furthermore, we also have the raw text of the Yelp Review as well as geographical information about the business and photos for some businesses, which opens the possibility of using moderns visual image recognition and natural language processing techniques to further extract node-level meta-data to incorporate into our model.

Concretely, we focus our work on predicting the rating that a user will assign a particular business. This problem has immediate and obvious utility: it would be useful to help users discover new businesses to visit (if the predicted rating is high) and also help business determine positive and negative trends. The dataset can be broken into three sets so we can train, evaluate, and test our models. One set will have edges, reviews, and information for businesses for a certain time $[t_0, t_1)$, the second set will have the edges created from $[t_1, t_2)$ and will be used to cross-validate our models and tune hyper-parameters, and the third set will he a hold out containing edges from $[t_2, t_3)$ and will be used for testing only.

\subsection{Network-Only Predictor}
We first present a predictive model which focus ``only'' on the structure of the graph, and uses this information to predict the ratings. For this purposes, we focus on the smaller user/business graph as shown in Figure \ref{fig:graph_structure}. We therefore have an undirected, weighed graph. In later sections, we explore alternative representations as well as additional data which can be input to our learning models.

\subsubsection{Data Preprocessing}
Given this representation $G$, we define three sets - training, validation, and test. We split the graph naturally -- edges and nodes are added to the graph as time progresses. However, we make special care to only use the nodes which remained and were available in the graph for the extent of our study. We can see the distribution of the reviews (edges) in our graph over time in Figure \ref{fig:reviews_over_time}. Given the skewed nature of the graph, we subset it to include only the latest reviews. Let us consider $G, G_{train}, G_{val}$ and $G_{test}$ where $G = G_{train} \cup G_{val} \cup G_{test}$. We first perform the following to obtain $G$:

\begin{figure}[h!]
\centering
\includegraphics[width=0.5\textwidth]{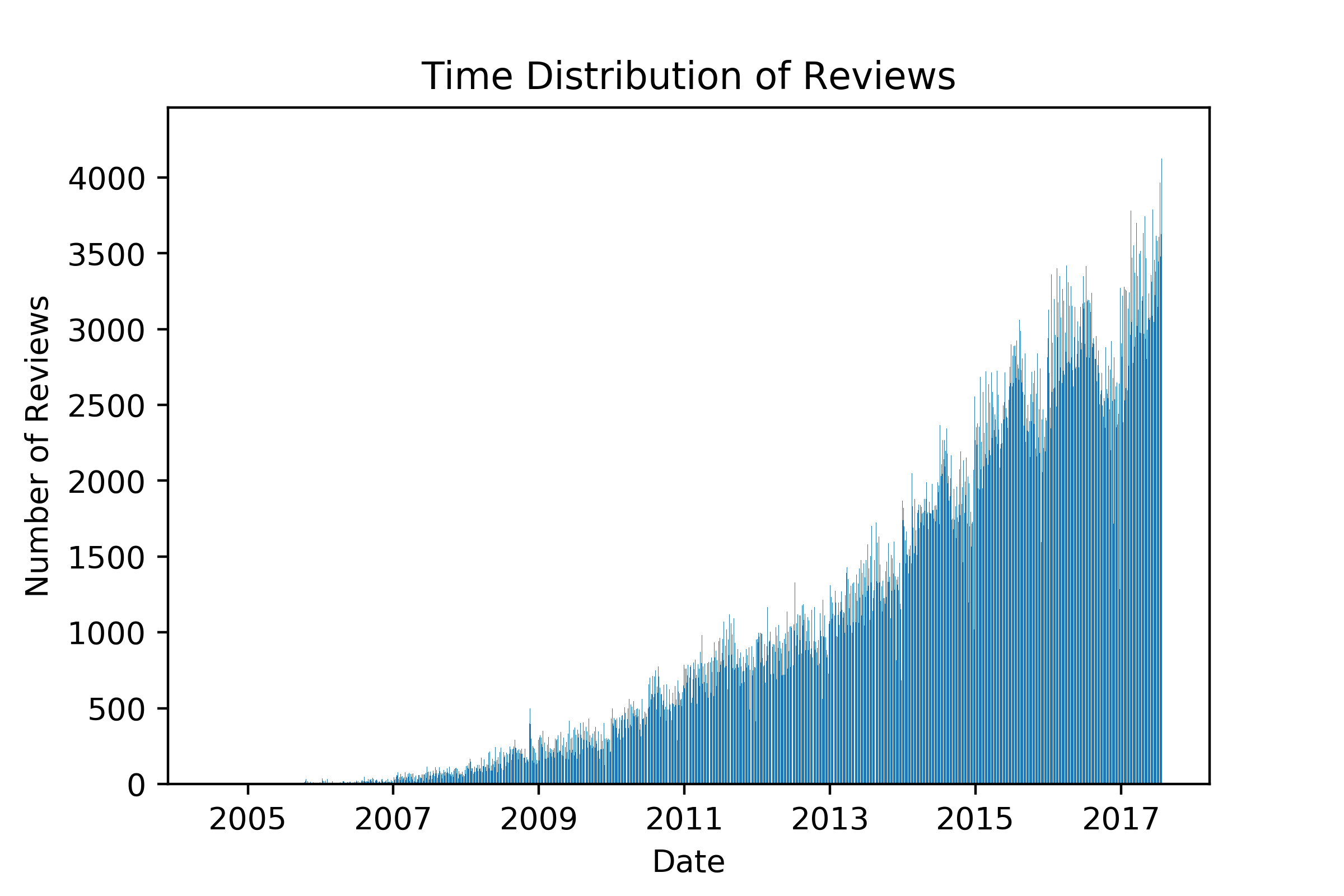}
\caption{Number of reviews in original dataset as a measure of time. We can see readily that the number of reviews increases drastically in the later years.}
\label{fig:reviews_over_time}
\end{figure}

\begin{figure}[h!]
\centering
\includegraphics[width=0.5\textwidth]{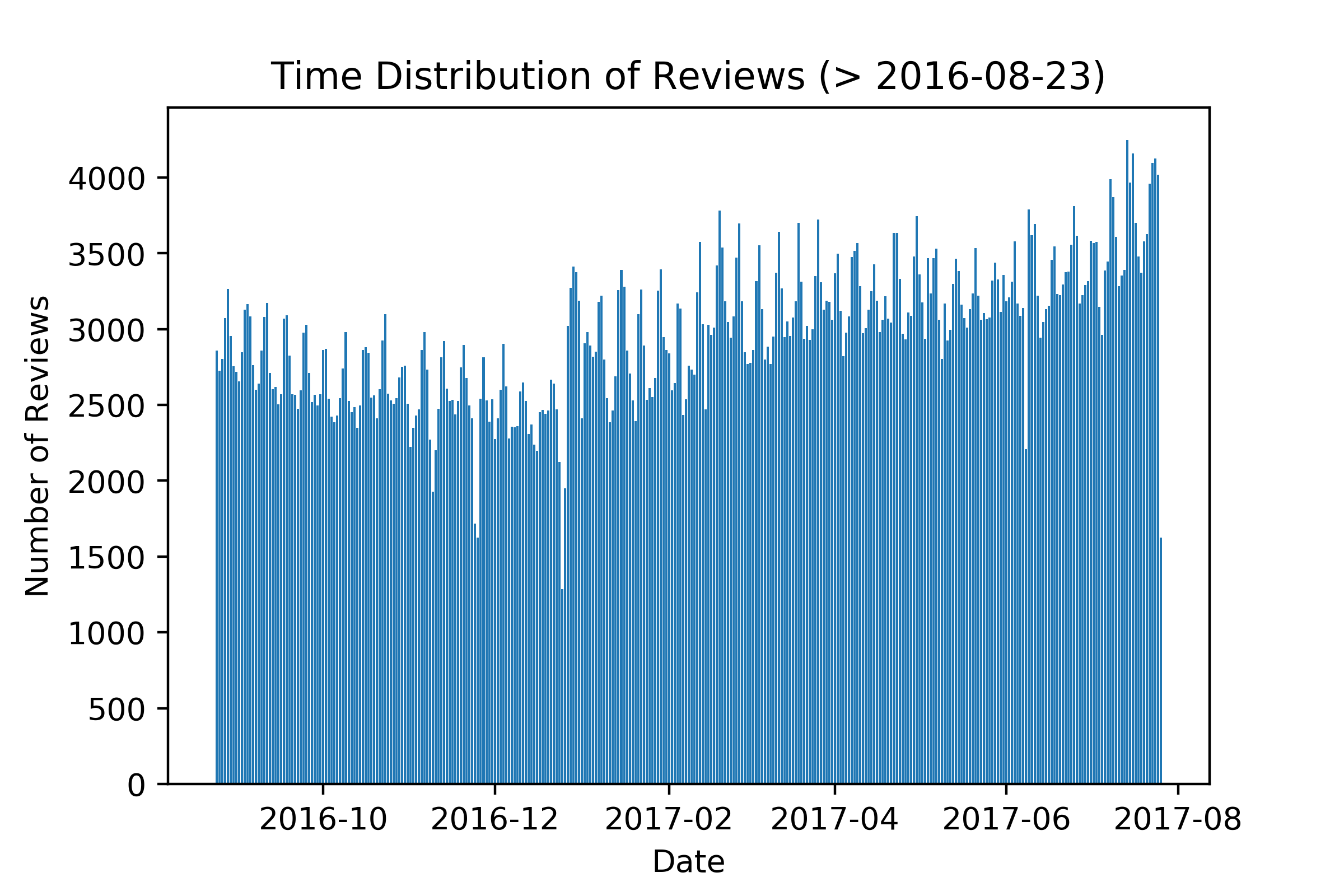}
\caption{Number of reviews by date for $G$}
\label{fig:reviews_subset_over_time}
\end{figure}

\begin{itemize}
\item Remove all reviews before ``2016-08-24''. This is primarily to (1) remove bias from early users and reviewers and instead focus on later reviews (see Figure \ref{fig:reviews_subset_over_time} for the distribution over time, which is far more uniform) and (2) reduce the size of our graphs to a manageable data set. We then have a graph with 428,795 users, 107,138 businesses, and 1,000,277 edges. We therefore have an extremely sparse graph, as only 0.0103109977941166\% of all possible edges even exist.
\end{itemize}

We now proceed to split the graph into $G_{train}, G_{test}$ and $G_{val}$. We split time-wise using three split-points, $t_0, t_1,$ and $t_2$. Then we have $G_{train}$ as the subset of $G$ from $[t_0, t_1)$ and $G_{val}$ as the subset in $[t_1, t_2)$ with $G_{test}$ containing the subset to the latest date $[t_2, \infty)$. Furthermore, we set all nodes in $G_{val}$ and $G_{test}$ to be the same set as those in $G_{train}$ to avoid running into issues with unseen nodes in the network.

After the above, we end up with the following networks:
\begin{itemize}
\item $G_{train} = (V_{train}, E_{train})$ with $|V_{train}| = 375,149$ where we have $283,085$ users and $92,064$. We also have $|E_{train}| = 599,133$, which is an incredibly sparse graph given (only $0.00143945169412\%$ of edges exists, even taking into account the bipartite structure of the graph).
\item $G_{val} = (V_{val}, E_{val})$ with $|V_{val}| = 75,466$ and $|E_{val}| = 88,079$.
\item $G_{test} = (V_{test}, E_{test})$ with $|V_{test}| = 67.125$ and $|E_{val}| = 73,730$.
\end{itemize}

Note that we've split the data essentially into an $80\%, 10\%, 10\%$ split. For more details on the graph structures, see Appendix \ref{sec:graph_distributions}. Given the sparsity of the graph, we focus on predicting the star rating given that an edge is created between user $u$ and business $b$. As such, our dataset does not contain any negative examples. We leave this predictive problem for open investigation. Furthermore, given the extreme size of our data, we process and train our models using Google Compute Engine with 8 CPUs and 30GB of memory.

\subsubsection{Graph Features}
Now that we have partitioned our data into training, validation, and testing, we move forward with calculating some rating prediction scores. We first focus on calculating multiple properties from our generated graph. In fact, we calculate the following:

\begin{itemize}
\item \textbf{Number of Common Raters:} For each pair $(u,b)$ of user and business, we calculate the number of common raters. A common rater is an extension of neighbors, in the sense that this is someone who has also rated $b$.
\item \textbf{Number of Common Business:} For each pair $(b,u)$ of user and business, we calculate the number of common businesses. A common business is an extension of neighbors, in the sense that this is a business someone who has also rated $b$.
\item \textbf{Average Rating of Common Raters:} For each pair $(u,b)$ of user and business, we calculate the number of common raters. A common rater is an extension of neighbors, in the sense that this is someone who has also rated $b$.
\item \textbf{Average Rating of Common Business:} For each pair $(u,b)$ of user and business, we calculate the number of common raters. A common rater is an extension of neighbors, in the sense that this is someone who has also rated $b$.
\item \textbf{Preferential Attachment}: We take the product of the average star rating of businesses rated by $u$ and the average star rating of raters of business $b$. We expect this value to indicate the relative popularity.
\item \textbf{Page Rank:} We treat the graph as an unweighed undirected graph and calculate the page rank value for all nodes and assign their sum as a feature.
\item \textbf{Eigenvector Centrality}: We calculate the global centrality of a node (compared to its neighbors) and use the sum as a feature.
\item \textbf{Adamic-Adar measure}: We look at common neighbors (as defined previously) and sum the inverse of the sum of their degrees (considering the graph to be weighed). Intuitively, this creates a measure for similarity where nodes with the same degreed neighbors are more similar.
\end{itemize}

Once calculate for our training, validation, and test data sets, all of the features are normalized to have unit mean and unit variance as is standard practice in machine learning problems.

\subsubsection{Models for Prediction}
We now present and describe the machine learning models used from the extracted features. Let $X$ be our training matrix, which is of shape $(n,d)$ where $n$ is the number of training examples and $d$ is the number of features extracted (in our case, $d = 9$) and $n = 599,133$. The most straight forward approach is simply to integrate our extracted features individually and directly train our models to predict the ratings, in a scale from $0$ to $5$. We now present the models we attempted.

\begin{enumerate}
\item \textbf{Linear Regression}: We attempt to fit a standard linear regressors to our input feature set. That is to say, our model takes the form of $r_i = \sum_{d=1}^{D} w_d x_{id}$ where $x_i$ is a single feature vector in our training set and $r_i$ is the corresponding rating. We train the model directly using the generated data from above and directly on the raw ratings for each edge. Linear regression is a simple model which attempts to minimize the mean square error, and can be thought of as a data generating process where we assume the ratings $r$ are generated by $r = W^TX + b + \epsilon$ where $\epsilon \sim N(0,\sigma)$ is some noise introduced into the system. The models is then able to recover the best plane of fit $W$ such that the error is minimized. In terms of loss functions, we can consider this as minimizing the loss function $L:$
\begin{align*}
L(W,b; X,y) &= \sum_{i=1}^{|X|} (\hat{y}_i - y_i)^2 \\
&=\sum_{i=1}^{|X|} (w_i^Tx_i + b_i - y_i)^2 \\
&=\sum_{i=1}^{|X|} \left(\sum_{d = 1}^{|x_i|} w_dx_{id} + b_i - y_i\right)^2
\end{align*}
where $x_i$ is a the $i$-th row in our feature matrix $X$. We minimize over the parameters $W$.
\item \textbf{Ridge Regression}: This is an improvement of linear regression. A possible issue with normal linear regression is that the possibility of over-training on the training set. It is possible to generate an extremely ``peaky'' set of weights such that the training error is reduced significantly yet the test error increases. The issue here is that we lack any term enforcing generalization in our loss function. The most typical method to enforce this generalization is to add a regularizer to the weights $W$. The loss function then becomes:
\begin{align*}
L(W, b; X,y) &= \sum_{i=1}^{|X|} (\hat{y}_i - y_i)^2 + \alpha|W|\\
&=\sum_{i=1}^{|X|} (w_i^Tx_i + b_i - y_i)^2 + \alpha \left|\sum_{i,j} W_{ij}^2\right|\\
&=\sum_{i=1}^{|X|} \left(\sum_{d = 1}^{|x_i|} w_dx_{id} + b_i - y_i\right)^2 + \alpha \left|\sum_{i,j} W_{ij}^2\right|
\end{align*}

The above encourages the model to minimize the squared loss for the training data while still maintaining a relatively sparse matrix $W$. This further prevent values in $W$ from becoming too large. In our case, we find $\alpha = 0.0001$ to be the optimal hyper-parameter (tuned on the validation set).
\item \textbf{Bayesian Regression}: Bayesian regression is essentially equivalent to ridge regression, but it is self-regularizing -- this means we do not need to choose an optimal parameter $\alpha.$ The theory behind Bayesian regression is to consider finding the parameters $W$ in our mode $y = WX$ which maximize the model probability. Given Bayes' rule, we have:

\begin{align*}
P(W,b \mid X) &= \frac{P(X \mid W,b)P(W,b)}{P(X)} \\
&\propto P(X \mid W,b)P(W,b) \\
\end{align*}
If we consider the case where $P(W,b) \sim N(\mu, \Sigma)$, then we arrive at ridge regression. We use this Bayesian model to also directly predict our ratings $r$. We optimize the above using the ADAM gradient descent optimizer where we use $\alpha_1 = \alpha_2 = \lambda = \lambda_2 = 0.000001$. The parameters are not tuned using the validation set due to lack of computational resources.

\item \textbf{Deep Neural Networks}: The latest research has had great success using ``deep learning'' to extract more details from the data and to learn the values and result more directly. We make use of this approach by constructing a relatively shallow network consists of a fully connected layer with 200 neurons, followed by a second fully-connected layer with 40 neurons, followed by a fully connected layer of 8 neurons, and a final fully connected layer of 2 neurons.

Given the recent effectiveness in a large range of tasks of this model, we expect that it will likewise be useful for rating prediction.

This gives us a total of $200x(9 + 1) + 40x200 + 8x40 + 2x8$ with a relu nonlinearity:
$$
relu(x) = \max(0,x)
$$
We use a final softmax at the end to generate the distribution of ratings.

We use Adam to perform gradient descent (with parameters $\beta_1 = 0.9$ and $\beta_2 = 0.999$ and $\epsilon = 1\times10^{-8}$) on the loss function with a regularization factor or $\alpha = 0.0001$ and batch size of $200$. We maintain a constant learning rate of $0.001$, randomly shuffle the input data. The parameters are selected based on past experience with neural network training and are not optimized using cross-validation or the validation set. 

\item \textbf{Random Forest}: We make use also of a random forest estimator. The random forest is a meta estimator that fits a number of classifying decision trees on various sub-samples of the dataset and uses averaging to improve the predictive accuracy and control over-fitting. We generate the sub-samples by sampling from the original data set with replacement. We use a total of 100 estimators, where we look at all features in the data set when considering the best split. 
\end{enumerate}

\subsection{Method Evaluation}

For all of the above approaches, we evaluate our effectiveness on the validation set and use this to tune our hyper-parameters (ie, network size, learning rate, etc.). In the end, we evaluate the results on the test set (previously unseen and untouched by our models) and make predictions for ratings in the seen edges. 

We evaluate our models across three metrics:

\begin{itemize}
\item The root mean squared error. This evaluates how close our predictions achieve our desired ratings:
$$
RMSE = \sqrt{\frac{1}{|N|}\sum_{i = 1}^{|X|} (\hat{y}_i - y_i)^2}
$$
\item The relative error. This is a metric that evaluates, on average, how wrong our star rating is compared to the true star rating. We take the method to be more indicate of improvements in our algorithms and with our data extraction. Formally, we define the relative error as:
$$
RELERROR = 100*\frac{1}{|X|}\sum_{i=1}^{|X|} \frac{|\hat{y_i} - y_i|}{\max_{i=1}^{|X|} \max\{\hat{y_i}, \hat{y}\}}
$$
\item The last metric we use for evaluating our regression models is the $R^2$ score. This gives us a way to evaluate our models against other models in literature, as is standard across regression problems. The best possible score is 1.0. The score ranges from $(-\infty, 1.0]$, where the worse a model is the more negative the value. Note that in the case where we have a model which simply predicts the constant expected value of the final output (disregarding any input features):
$$
E[Y] = \frac{1}{|X|}\sum_{i=1}^{|X|} y_{i}
$$
we will have a score of $0.0$. The formula for computing this score is:
$$
R^2 = 1 - \frac{\sum_{i=1}^{|X|}(y_i - \hat{y}_i)^2}{\sum_{i=1}^{|X|} \left(y_i - \frac{1}{|X|}\sum_{i=1}^{|X|}y_i\right)^2}
$$

\end{itemize}

\subsection{Extracting Item Features}

Given our results (see Results section) from the above models, we continue forward with our deep neural network. We begin by augmenting the data available for the business nodes. 

\begin{figure}[h!]
\centering
\includegraphics[width=0.5\textwidth]{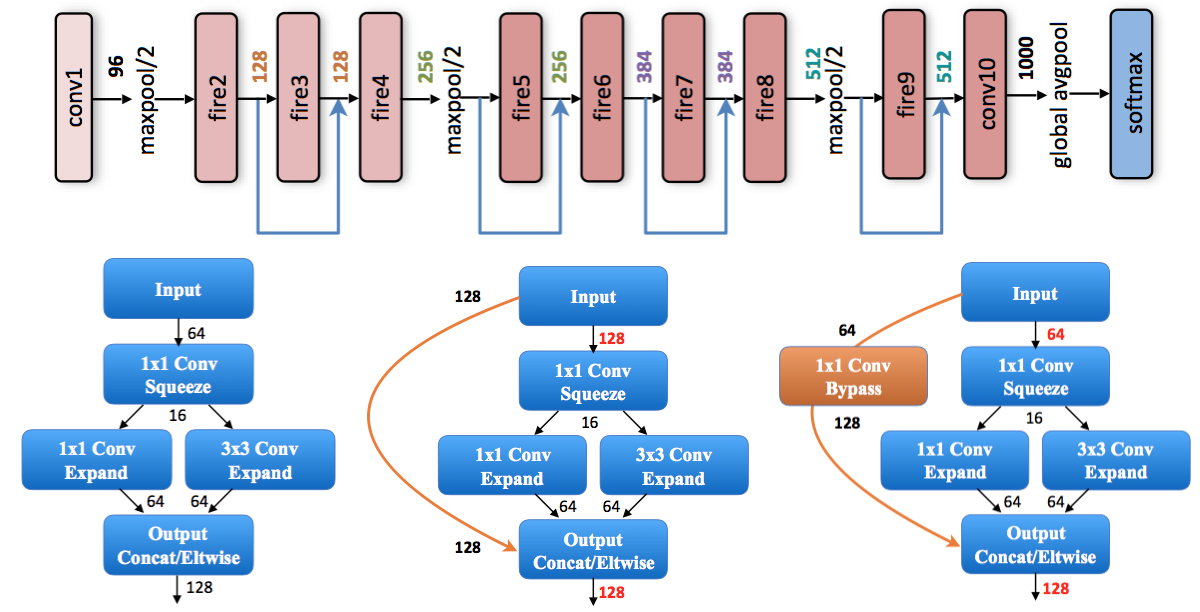}
\caption{Original SqueezeNet Architecture. We modify it to remove the final soft-max layer and instead output a $1000$ embedding for our images.}
\label{fig:squeezenet_architecture}
\end{figure}

\begin{itemize}
\item We make use of the pre-trained SqueezeNet network included in the PyTorch model zoo for visual image processing. We first down-sample the images to the expected 256x256x3 input (we do this simply by cropping and averaging pixel values over regions mapping into the 256x256xspace).
\item We can then feed these smaller images directly into the pre-trained squeezenet (see Figure \ref{fig:squeezenet_architecture} for architecture) which has been modified to remove the final soft-max layer (and instead we produce a vector in $\mathbb{R}^1000$.
\item For a business $b$, we take the $p_i^b \in \mathbb{R}^1000$ and compute their mean. We use this embedding as a representation of the business.

\item Furthermore, we make use of the pre-trained word-embedding and take the business description and generate a small 256-dimensional vector.
\item We concatenate the above vectors into a 1000 + 256 + 9 vector, which we take as input into a modified neural net.
\end{itemize}

The meat of the model consists of a neural net which takes a input as 1265-length vector for each $(u,b)$ pair and runs through through a single layer with 200 hidden units (so number of parameters is 200x1265). We then take this and feed it into the successful network described in the previous section and evaluate it in the same way as described before.

\subsubsection{Training and Date}

Due to the large size of the above networks, we subset the data significantly into a much smaller amount of only ~15k reviews. We select the businesses with the most photos as the candidates to subset by, and make sure we take the reviews which include these businesses. With the reduced data size, we are able to successfully train our specified model end-to-end and achieve a marginal improvement over our previous models.

\section{Results and Discussion}
In this section, we presents the results on our test set and provide some discussion as to the extent to which our models successfully predicted Yelp review ratings.

{\renewcommand{\arraystretch}{2}%
\begin{table*}[]
\centering
\caption{Supervised Training Results on Training Set}
\label{table:trainint_set_results}
\begin{tabular}{|l|lll}
\hline
\textit{\textbf{Model}}      & \multicolumn{1}{l|}{\textbf{RMSE}} & \multicolumn{1}{l|}{\textbf{RELERROR}} & \multicolumn{1}{l|}{\textbf{R\textasciicircum 2}} \\ \hline
\textit{Baseline}            & 1.50142049076                      & 25.7312431633                          & 0.0                                               \\ \cline{1-1}
\textit{Linear Regression}   & 1.29409210615                      & 20.397681968                           & 0.257107970487                                    \\ \cline{1-1}
\textit{Ridge Regression}    & 1.29409210617                      & 20.3976788744                          & 0.257107970462                                    \\ \cline{1-1}
\textit{Bayesian Regression} & 1.29409213097                      & 20.3975770983                          & 0.257107941987                                    \\ \cline{1-1}
\textit{Neural Network}      & 1.26509191767                      & 18.7831282852                          & 0.290030838364                                    \\ \cline{1-1}
\textit{Random Forest}       & \textbf{0.749654164334}            & \textbf{10.0184313324}                 & \textbf{0.750702893173}                           \\ \cline{1-1}
\textit{\textbf{Business Features}}       & 1.24943163247            & 16.2852635532                 & 0.32123445344                           \\ \cline{1-1}
\end{tabular}
\end{table*}
}


{\renewcommand{\arraystretch}{2}%
\begin{table*}[t]
\centering
\caption{Supervised Training Results on Validation Set}
{}\label{table:validation_set_results}
\begin{tabular}{|l|lll}
\hline
\textit{\textbf{Model}}      & \multicolumn{1}{l|}{\textbf{RMSE}} & \multicolumn{1}{l|}{\textbf{RELERROR}} & \multicolumn{1}{l|}{\textbf{R\textasciicircum 2}} \\ \hline
\textit{Baseline}            & 1.42776997765                      & 24.1191750901                          & 0.0                                               \\ \cline{1-1}
\textit{Linear Regression}   & 1.18380441761                      & 18.2853412809                          & 0.327892972837                                    \\ \cline{1-1}
\textit{Ridge Regression}    & 1.18380391943                      & 18.2853539128                          & 0.312515072251                                    \\ \cline{1-1}
\textit{Bayesian Regression} & 1.18378759889                      & 18.2857790551                          & 0.312534028173                                    \\ \cline{1-1}
\textit{Neural Network}      & \textbf{1.16442192777}             & \textbf{16.0869919882}                 & \textbf{0.334842664919}                           \\ \cline{1-1}
\textit{Random Forest}       & 1.18801209881                      & 18.6598898159                          & 0.307618649842                                    \\ \cline{1-1}
\textit{\textbf{Business Features}}       & 1.14952444234            & 14.8854451849                 & 0.35986245424                           \\ \cline{1-1}
\end{tabular}
\end{table*}
}
{\renewcommand{\arraystretch}{2}%
\begin{table*}[t]
\centering
\caption{Supervised Training Results on Test Set}
\label{table:test_set_results}
\begin{tabular}{|l|lll}
\hline
\textit{\textbf{Model}}      & \multicolumn{1}{l|}{\textbf{RMSE}} & \multicolumn{1}{l|}{\textbf{RELERROR}} & \multicolumn{1}{l|}{\textbf{R\textasciicircum 2}} \\ \hline
\textit{Baseline}            & 1.4634860104                       & 24.7465614817                          & -0.000855120695457                                \\ \cline{1-1}
\textit{Linear Regression}   & 1.19928440313                      & 18.5669587686                          & 0.327892972837                                    \\ \cline{1-1}
\textit{Ridge Regression}    & 1.19928405085                      & 18.5669755633                          & 0.327893367682                                    \\ \cline{1-1}
\textit{Bayesian Regression} & 1.19927256299                      & 18.567542783                           & 0.327906243749                                    \\ \cline{1-1}
\textit{Neural Network}      & \textbf{1.1838237529}              & \textbf{16.3219078547}                 & \textbf{0.34511029369}                            \\ \cline{1-1}
\textit{Random Forest}       & 1.19281377927                      & 18.7137709175                          & 0.335125985417                                    \\ \cline{1-1}
\textit{\textbf{Business Features}}       & 1.1694425252            & 15.4556500245                 & 0.35111454552                           \\ \cline{1-1}
\end{tabular}
\end{table*}
}

We now present the final results from our models, each evaluated on the test set. We compare the different methods used, and discuss their differences and possible improvements. The main results are presented in Table \ref{table:test_set_results}, with the validation data set results in Table \ref{table:validation_set_results} and the training data set results in Table \ref{table:trainint_set_results}. 

We have implemented a complete end-to-end pipeline which begins with the (1) raw Yelp JSON data, (2) construct training, validation, and testing graphs over the data by our pre-determined timescales, (3) extracts training, validation, and testing sets from the graphs generated and computes a variety of network properties to be used by our machine learning models and (4) trains a variety of machine learning models on the extracted data and tunes their hyper-parameters when possible, (5) culminating in the evaluation of the models on the known results from the test set. We implements this work, wrote optimized code for feature extraction, and built our networks. Everything was implemented ourselves with the use of SNAP, Python, scikit-learn, and PyTorch -- the code base is publicly available at GitHub \footnote{https://github.com/kandluis/cs224w-project}.

The major challenge faced when experimenting was the sheer size of the dataset -- even after sub-setting the data to a more manageable size in the millions and using the extremely powerful Google Compute Engine to add additional memory and processing power, more complex models such as random forest and the convolution neural networks could take in the order of days to fully train. Even extracting the word embeddings and pre-trained image feature vectors with SqueezeNet and ResNet would alone take a significant amount of time -- so much so that it proved unfeasible to do for a large portion of the dataset.

As such, as described in our Methods section, we were able to sample only approximately one years worth of data from the Yelp review network. However, despite this, we were nonetheless able to train the network predictors on over 500k training samples (reviews) which contained over 280k users and over 92k businesses (for more than 375k nodes), and validate and test our network models on over 88K and 73K examples respectively.

Finally, to evaluate the performance of all of our models we make use of RMSE, relative error, and the score function defined in our methods. Our results can be see in Table \ref{table:trainint_set_results}, Table \ref{table:test_set_results}, and Table \ref{table:validation_set_results}.

\section{Discussion}
We begin the discussion by analyzing some network properties.

\subsection{Summary Statistics}
\subsubsection{Users}
\begin{figure}[h!]
\centering
\includegraphics[width=0.5\textwidth]{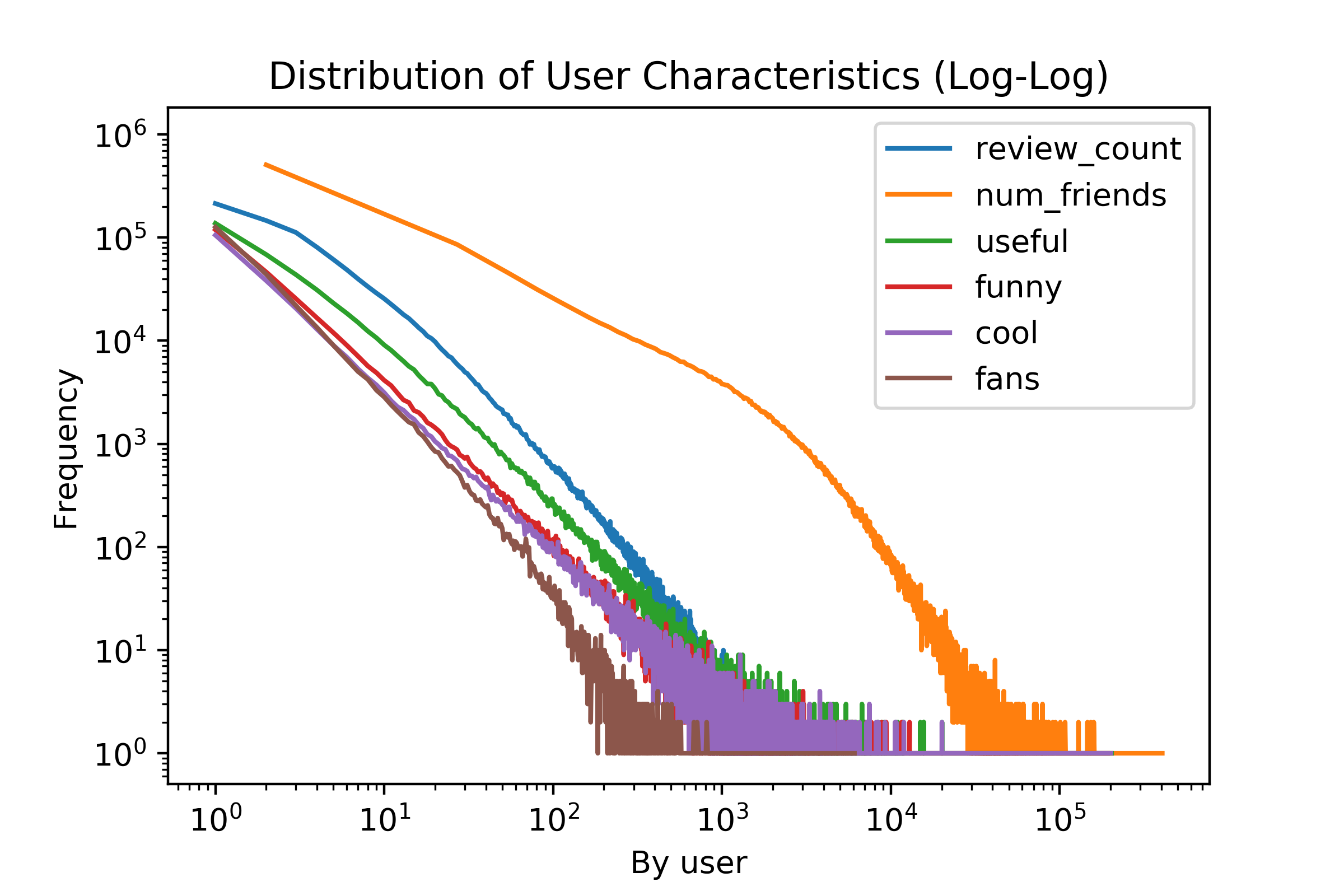}
\caption{Frequency of countable user characteristics -- the majority exhibit a power-law distributions}
\label{fig:user_characteristics}
\end{figure}

\begin{figure}[h!]
\centering
\includegraphics[width=0.5\textwidth]{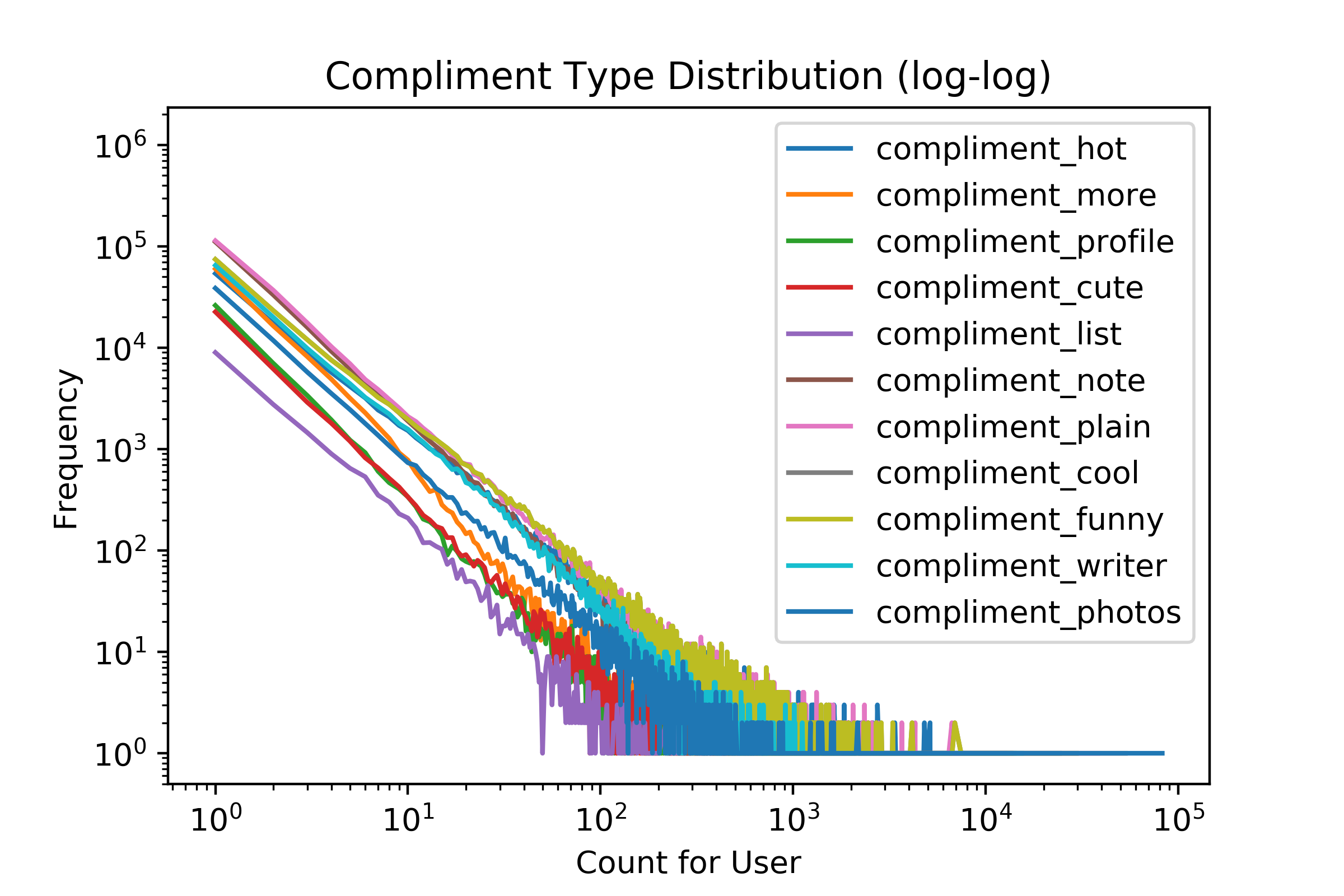}
\caption{Distribution of Received user Compliments}
\label{fig:user_compliment_distribution}
\end{figure}

We present some overview of the user meta-data. In Figure \ref{fig:user_characteristics}, we can see that multiple characteristics of the users follow a power-law degree distributions -- and not just the node degrees. This is to say that the distribution can be modeled as $P(k) \propto k^{-\gamma}$. The power-law distribution is immediately evident in:

\begin{itemize}
\item Number of review -- we have a few users that write many reviews and many users that write few reviews.
\item Number of friends -- this means the network follows a true-power law distribution.
\item useful/funny/cool/fans -- this appears to demonstrate that social ranking/status also follows a power-law distribution in the Yelp social network. This trend is further demonstrated by Figure \ref{fig:user_compliment_distribution}.
\end{itemize} 

\begin{figure}[h!]
\centering
\includegraphics[width=0.5\textwidth]{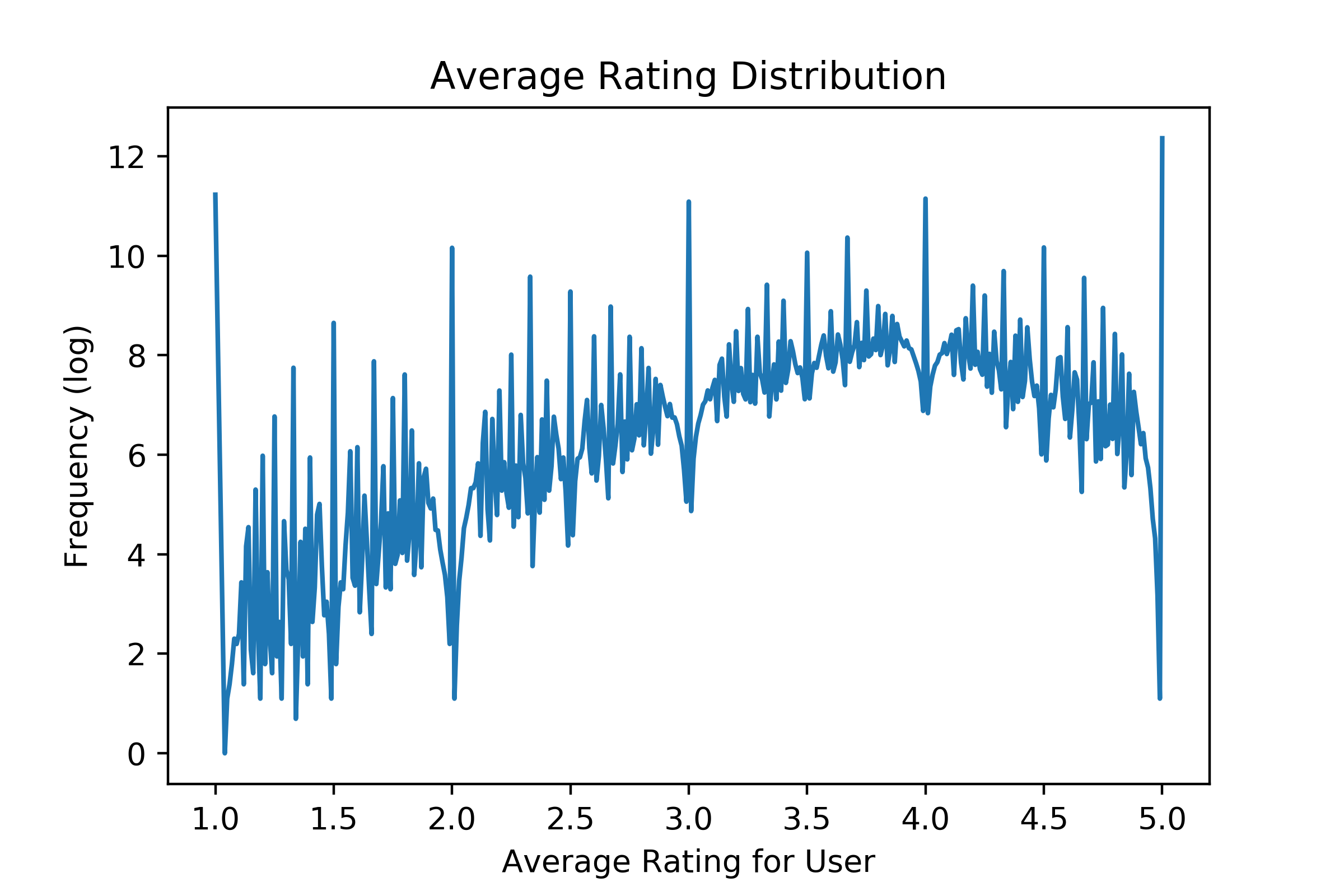}
\caption{Distribution of Average user Rating}
\label{fig:user_rating_distribution}
\end{figure}

Furthermore, we can look at the average rating given by users, across the network. The results are shown in a log plot in Figure \ref{fig:user_rating_distribution}.

We notice that the ratings tend to be inflated (3-5) stars being quite frequent, while 1-2 stars being very infrequent. Presumable this might be due to the fact that people do not frequent poor restaurants. The other aspect that is immediately apparent is the spikes at even numbered ratings -- this is likely due to users who have rated only one once, of which we have many.

\subsubsection{Businesses}
\begin{figure}[h!]
\centering
\includegraphics[width=0.5\textwidth]{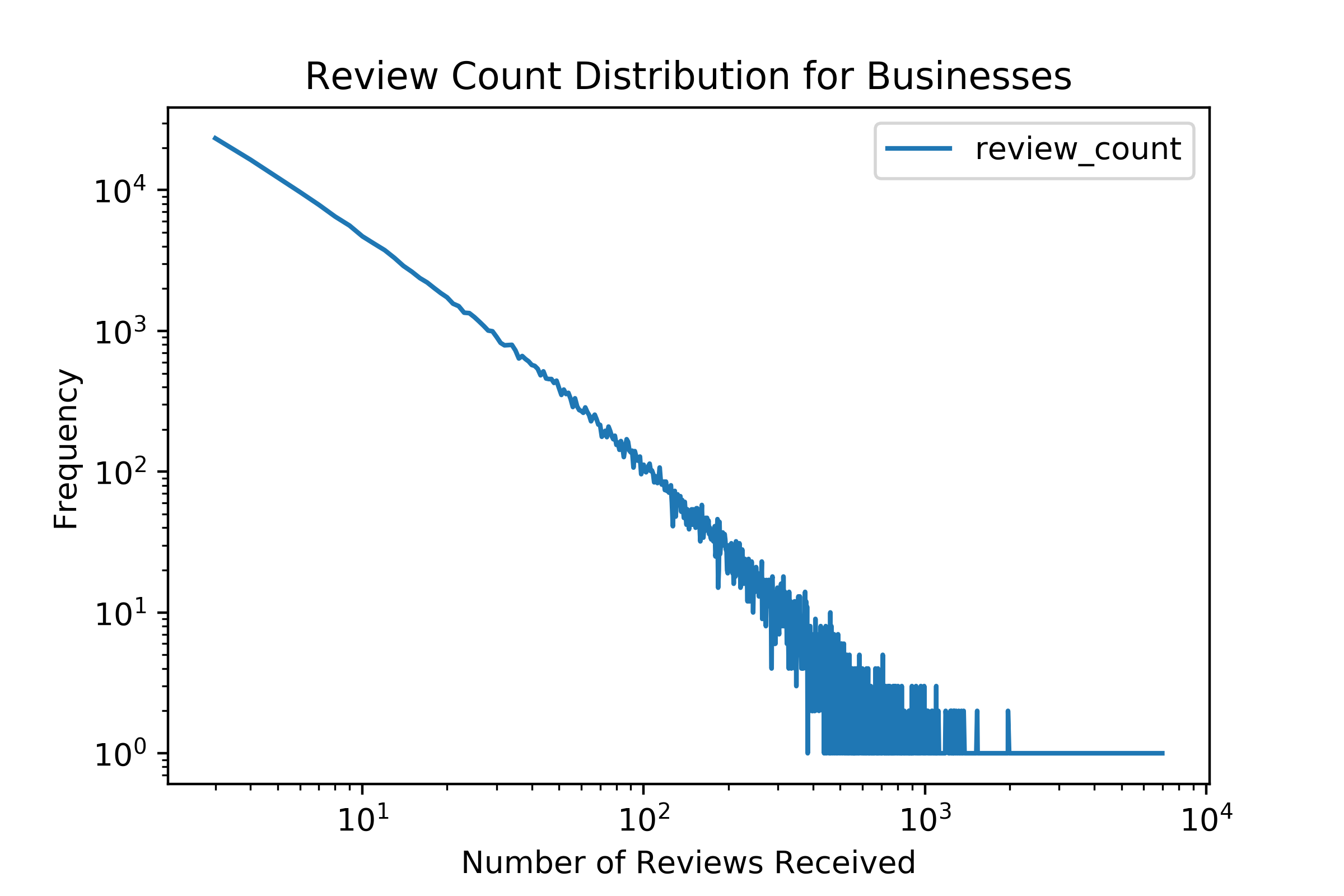}
\caption{Business Review Distribution}
\label{fig:business_review_distribution}
\end{figure}

\begin{figure}[h!]
\centering
\includegraphics[width=0.5\textwidth]{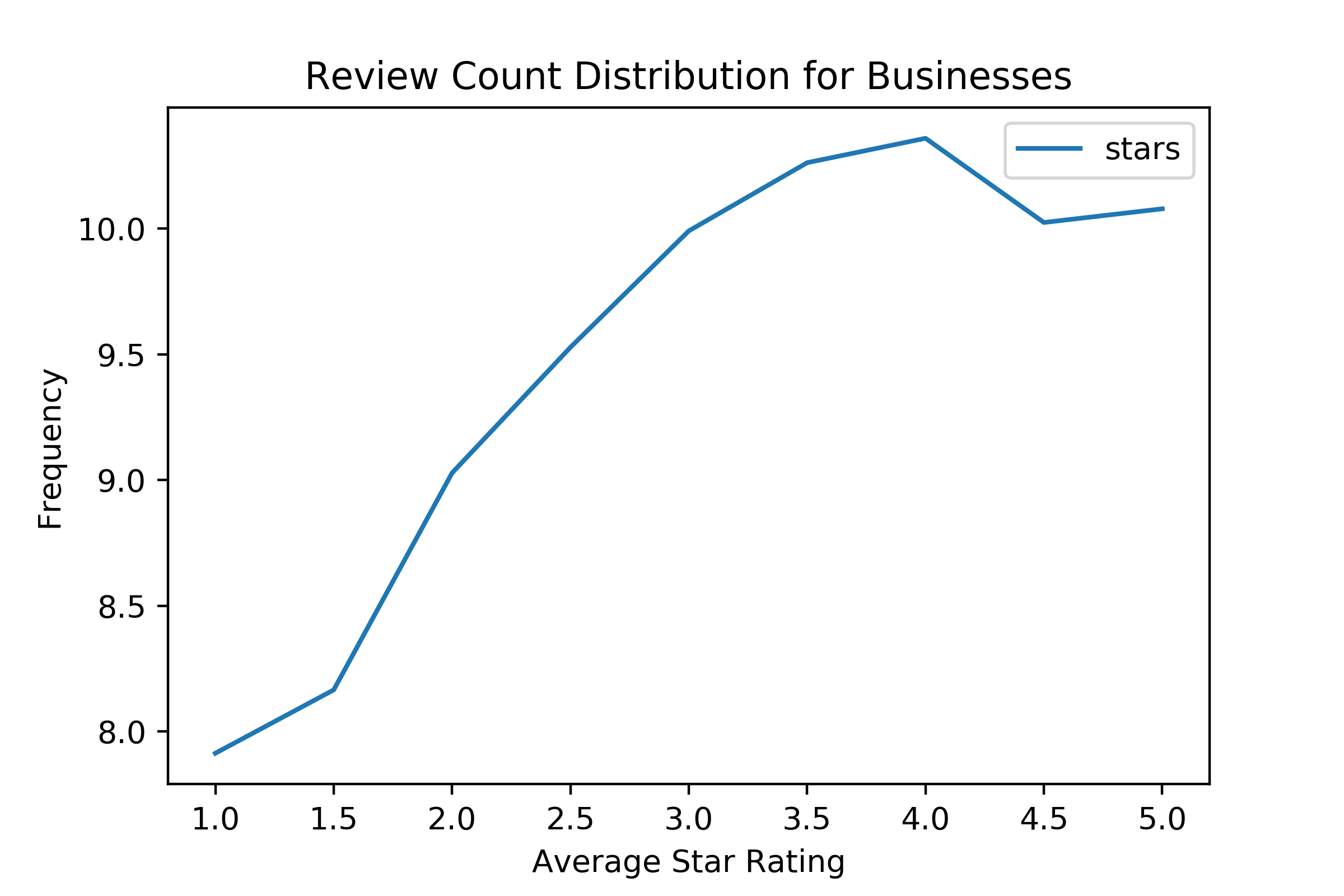}
\caption{Business Average Star Distribution}
\label{fig:business_star_distribution}
\end{figure}

We present some overview of the user meta-data. In Figure \ref{fig:business_review_distribution}, we can see that the power-law distribution is also respected in the business side. Furthermore, we can also see that businesses tend to be rated quite highly, on average, with most businesses either having in the range from 3-5 (see Figure \ref{fig:business_star_distribution}).

\subsection{Review Prediction}
We continue our discussing by now focusing on the results for our model predictions.

We can see that extracting network-only data and using machine learning models to fit the ratings seems to perform relatively well, even with simple, un-regularized linear regression. It appears that the features we selected, for example the nearest neighbors and the average ratings, are quite effective at both capturing network properties as well as capturing the user ratings. We did not need to extend the network to include further metadata information, and the results were nonetheless quite good, especially when compared to our non-trivial baseline.

\begin{figure}[h!]
\centering
\includegraphics[width=0.5\textwidth]{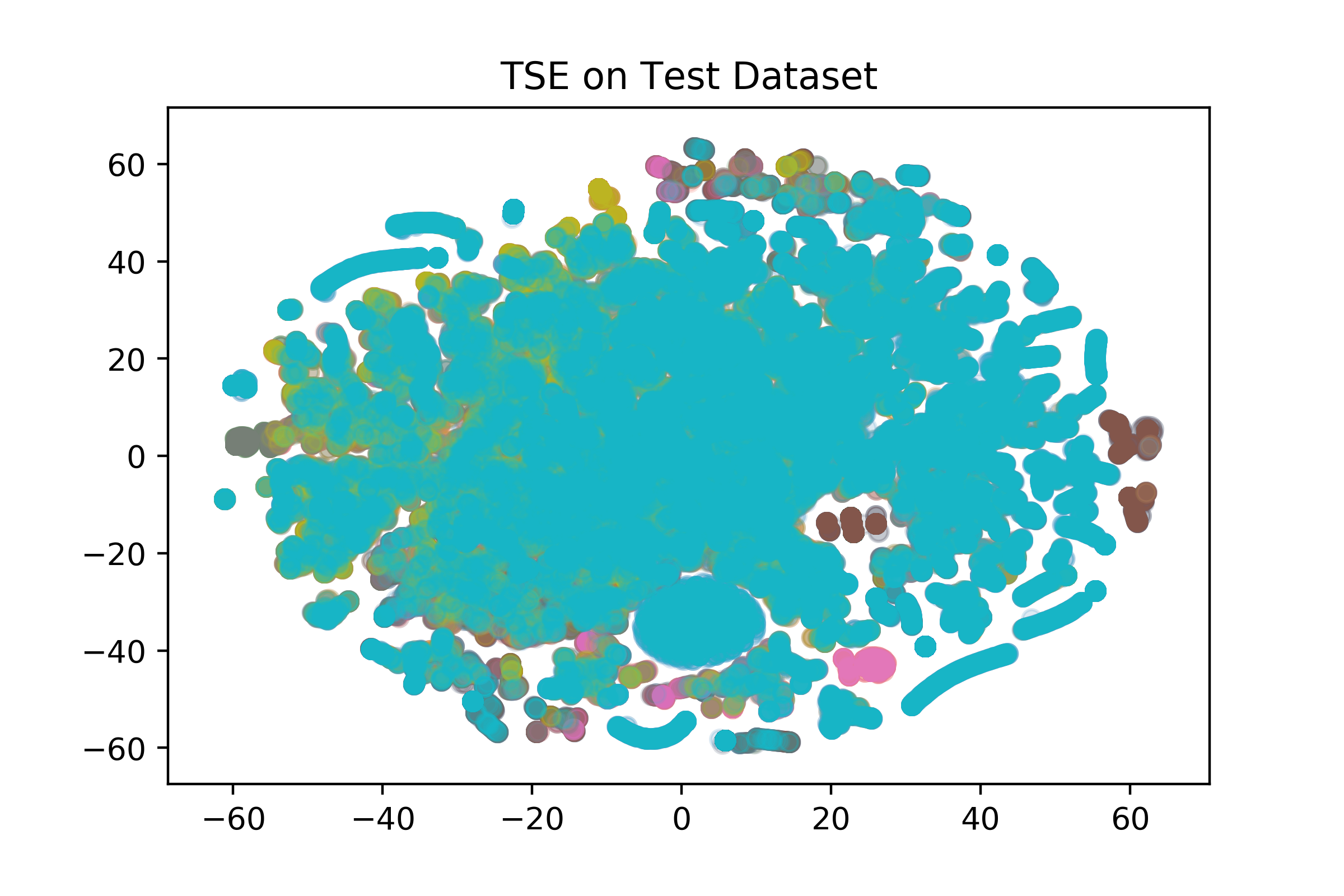}
\caption{t-SNE embedding of of $G_{test}$}
\label{fig:tsne_embedding}
\end{figure}

Furthermore, we note that our feature extraction proved extremely effective at generalizing across models. We see that in particular, the deep neural network and the random forest models both performed extremely well. It's interesting to note that the random forest model appears to have over-fit the data by a significant margin. This appeared promising on the training set but did not pan-out when we took the model to unseen data. However, we note that the neural net performed the best -- this appears to lead credence to the idea that the function learned is inherently non-linear, at least in the feature space we selected. This is somewhat counter to what we originally hypothesized, since all of the original features are approximately in the same scale as the ratings and would, intuitively, appear to predict the ratings rather directly. This idea is supported by the t-SNE embedding in Figure \ref{fig:tsne_embedding}, where we embed our test set in a lower dimensional space (given the features extracted) and we color each based on the rating given (from $1,2,3,4,5$). 

This intuitively gives us a good foundation for why the features we choose appear to be so well correlated. Furthermore, we note that the embedding shows a clear non-linear distribution, which appears to corroborate the results where our neural network performed the best. This seems to imply that a neural network would be the best approach to disambiguate between the possible ratings a user might assign to a business. We found the issue of predicting the lack of edges to be somewhat more nuanced and subtle, though initial experiments with this approach proved promising.

Another promising aspect involves using the photographic and textual descriptions of businesses as input features to our predictive models. Despite making use of pre-trained word-embeddings and pre-trained image models, the computational cost for the models proved extreme for our dataset. Sub-setting into smaller set showed some initially results which appeared positive, however the smaller dataset makes it difficult to determine whether the additional information was accurately fed into the models and used effectively. However, it does appear that the networks performed well overall, and at least learned to make use of the features from the images.

\section{Conclusion}
In this paper, we have presented a novel approach towards review rating prediction in a subset of the Yelp review network. We have investigated the effectiveness of network-only machine learning models, determining 9 key structural features of the network that proved effective in predicting the weight of unseen edges after using supervised algorithms to generate the models. We demonstrated that a deep neural net even with the limited feature set was the most effective and most general approach. Furthermore, we performed early experiments in making use of none-network features to improve the predictions of the neural network. We did this by creating a pipeline building on previous work were for each $(u,b)$ pair, the business descriptions were converted into their respective word-embeddings using the popular word2vect network followed by an RNN which output a fixed sized 256 feature vector for each business. Furthermore, we selected key images from the business and the photo dataset provided by yelp and ran them through pre-trained SqueezeNet network with the final classification layer removed to generate multiple 4096-dimensional feature vectors per image. These feature vectors were then averaged and fed as additional input into a final fully-connected neural network. These preliminary results showed marginal improvement in the accuracy of the results. This shows not only that our original models are able to understand higher order relationships within the Yelp review network, but are also able to understand features (and build on them) specific to each node.

\section{Future Work}
The project could be continued in several directions. In particular, we could continue to follow the example set by \cite{PintrestProject} and consider some of the temporal features of the graph structure. They proposed using a sliding window approach to achieve improved accuracy in link-prediction, which could easily be modified to support review prediction in the Yelp network.

Furthermore, our preliminary work incorporating deep convolution neural nets and recurrent neural nets to extract feature embeddings for the businesses have demonstrated marginal capabilities of improving the predictive power of our models. Further work could be done in this area by, rather than extracting static embeddings, incorporating the visual and textual networks into an end-to-end model which could tweak the learned weights for visual and textual processing in order to have better understanding of how these features related to the ratings given to businesses by users. Furthermore, would like to see further work placed into whether user features can similar be used to improve performance -- for example, finding embeddings of users based on their features and using these embeddings as inputs to our model.

Lastly, there's additional work to be done to incorporate even more graph features into the predictive model. Given the effectiveness of the network structure itself at predicting the values of unseen ratings alone, we would like to explore further network features and models and see how this additional information can improve our models. This can include incorporating the information about tips -- we would expect someone that has given a tip to be more likely to rate the business positively (or negatively).

In any case, we believe that there is yet much work to be done in this field and many potential interesting developments in the area of combining non-network features with network features.

\section{Acknowledgments}
The author would like to thank Jure Leskovec and the rest of the CS224W staff for their continued guidance and advice during the course of the project. In particular, for emphasizing a focus on network properties.

\bibliography{references}{}
\bibliographystyle{plain}

\section{Appendix}

\subsection{Graph Distributions}
\label{sec:graph_distributions}
\begin{figure}[h!]
\centering
\includegraphics[width=0.5\textwidth]{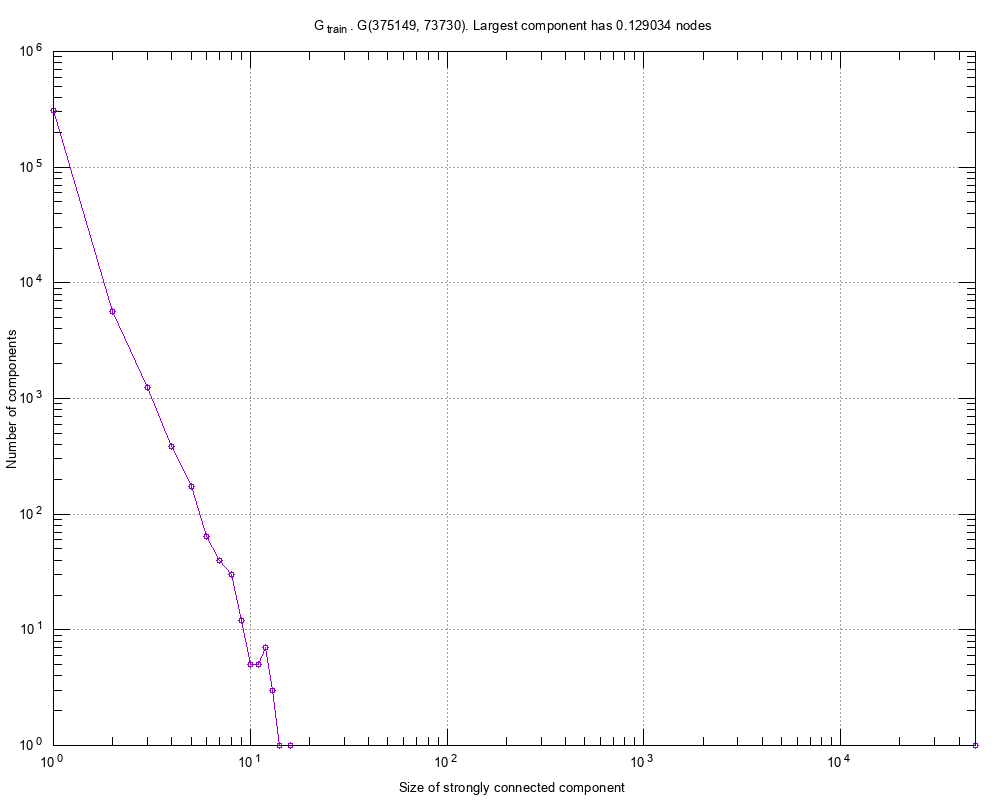}
\caption{Distribution of sizes of strongly connected components of $G_{train}$}
\label{fig:distribution_scc_g_train}
\end{figure}
\begin{figure}[h!]
\centering
\includegraphics[width=0.5\textwidth]{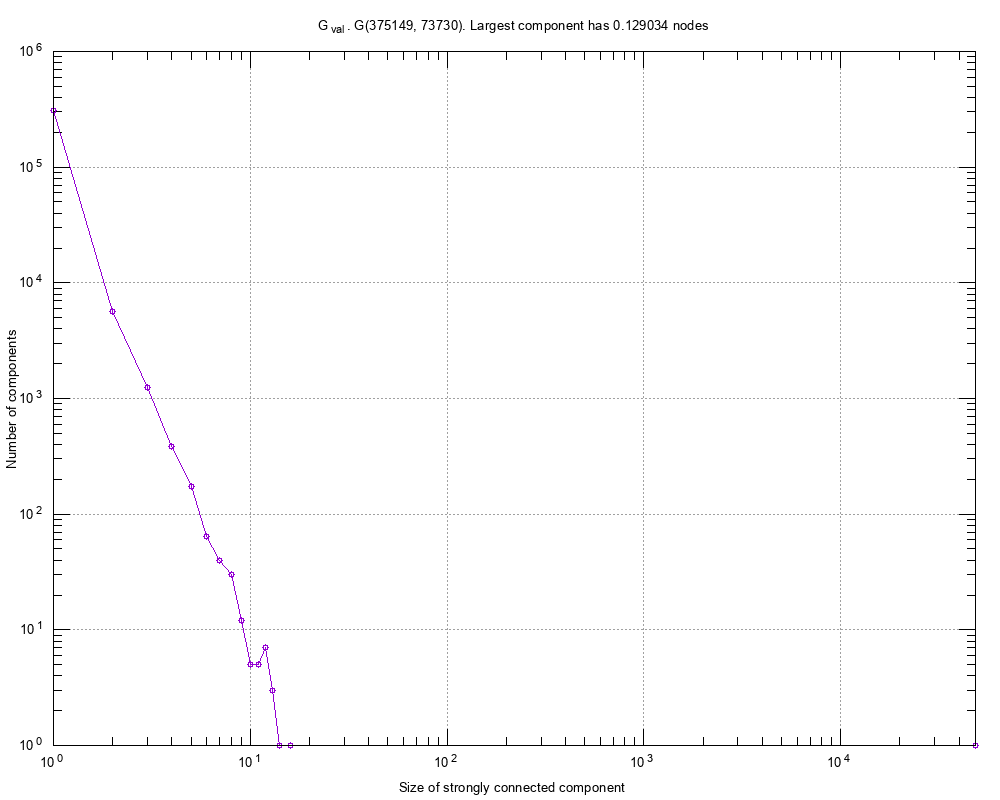}
\caption{Distribution of sizes of strongly connected components of $G_{val}$}
\label{fig:distribution_scc_g_val}
\end{figure}
\begin{figure}[h!]
\centering
\includegraphics[width=0.5\textwidth]{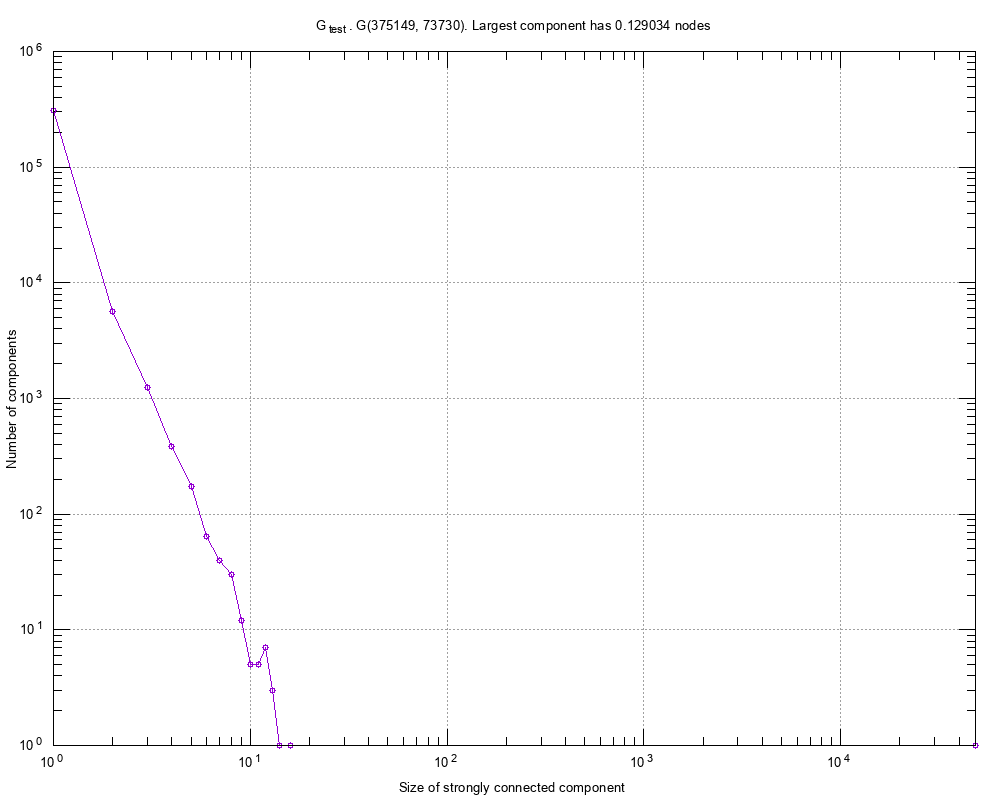}
\caption{Distribution of sizes of strongly connected components of $G_{test}$}
\label{fig:distribution_scc_g_test}
\end{figure}

\begin{figure}[h!]
\centering
\includegraphics[width=0.5\textwidth]{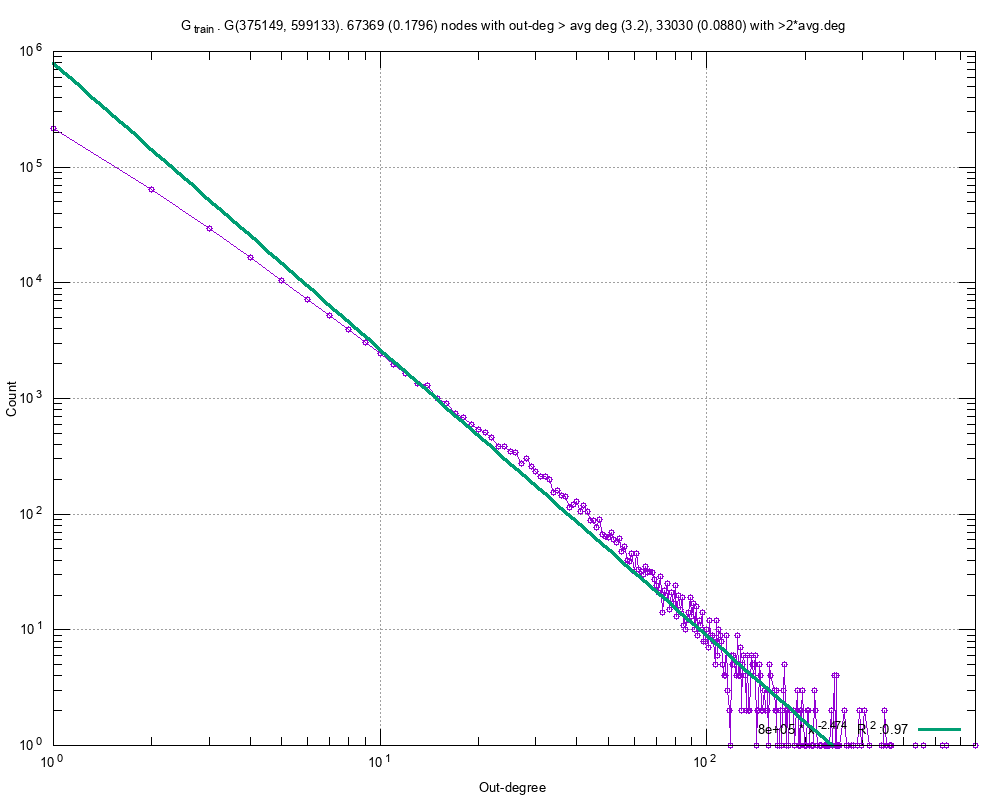}
\caption{Distribution of sizes of strongly connected components of $G_{train}$}
\label{fig:degree_distribution_g_train}
\end{figure}
\begin{figure}[h!]
\centering
\includegraphics[width=0.5\textwidth]{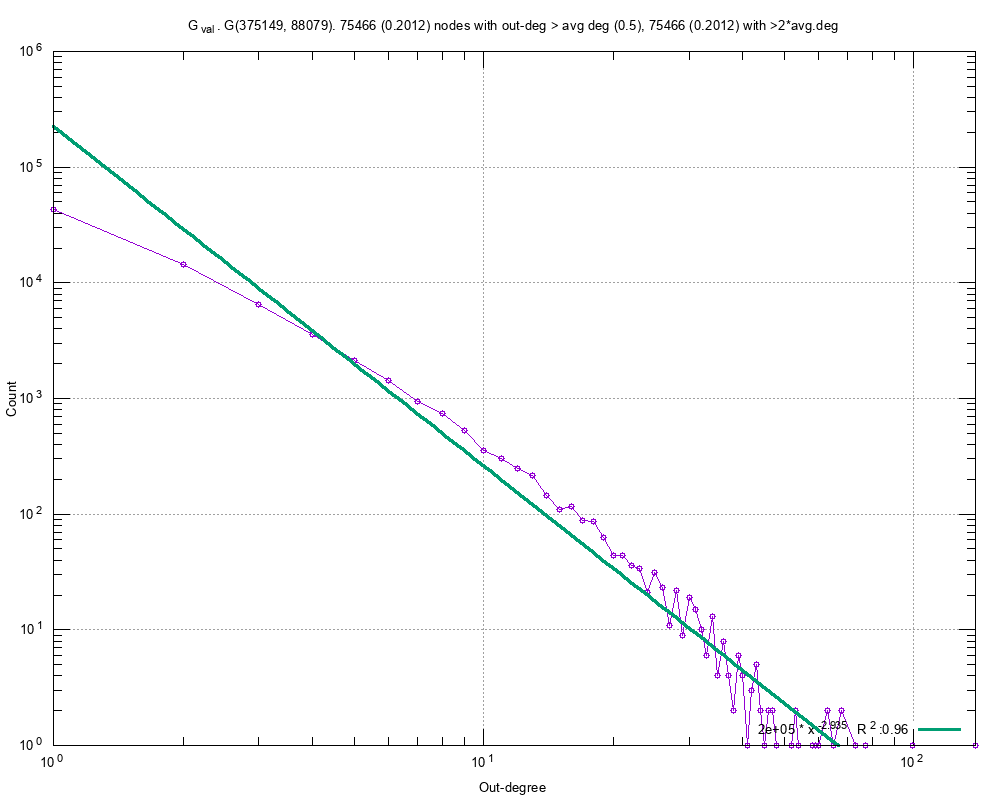}
\caption{Distribution of sizes of strongly connected components of $G_{val}$}
\label{fig:degree_distribution_g_val}
\end{figure}
\begin{figure}[h!]
\centering
\includegraphics[width=0.5\textwidth]{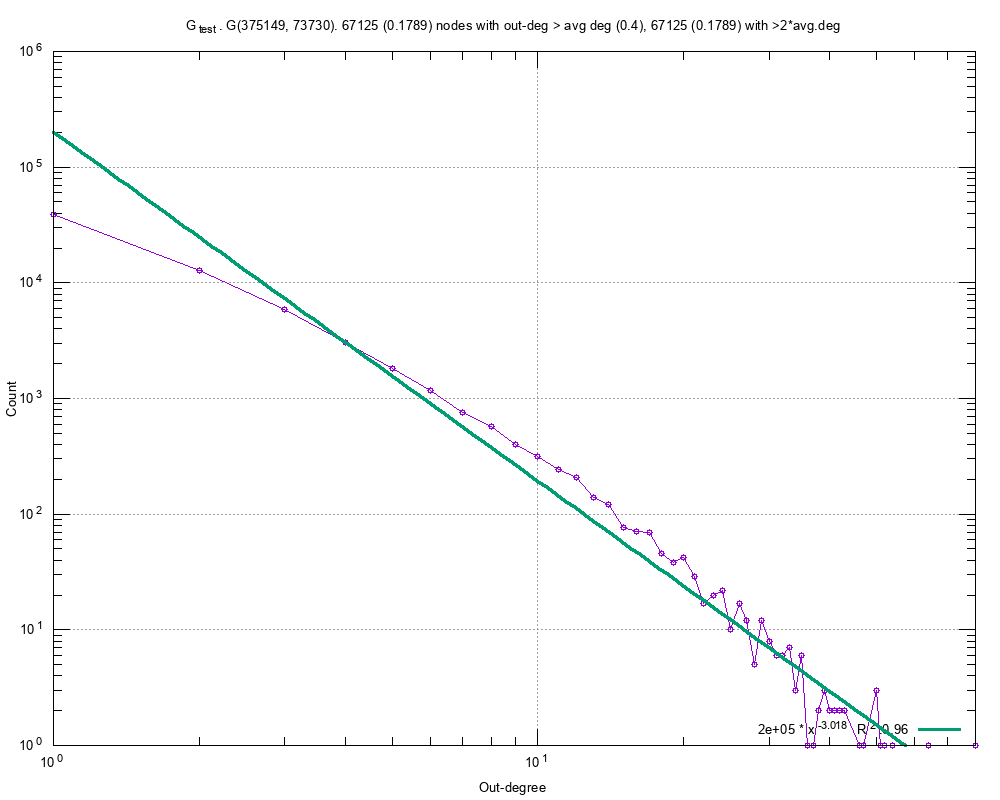}
\caption{Distribution of sizes of strongly connected components of $G_{test}$}
\label{fig:degree_distribution_g_test}
\end{figure}

\subsection{Other}

\begin{figure}[h!]
\centering
\includegraphics[width=0.5\textwidth]{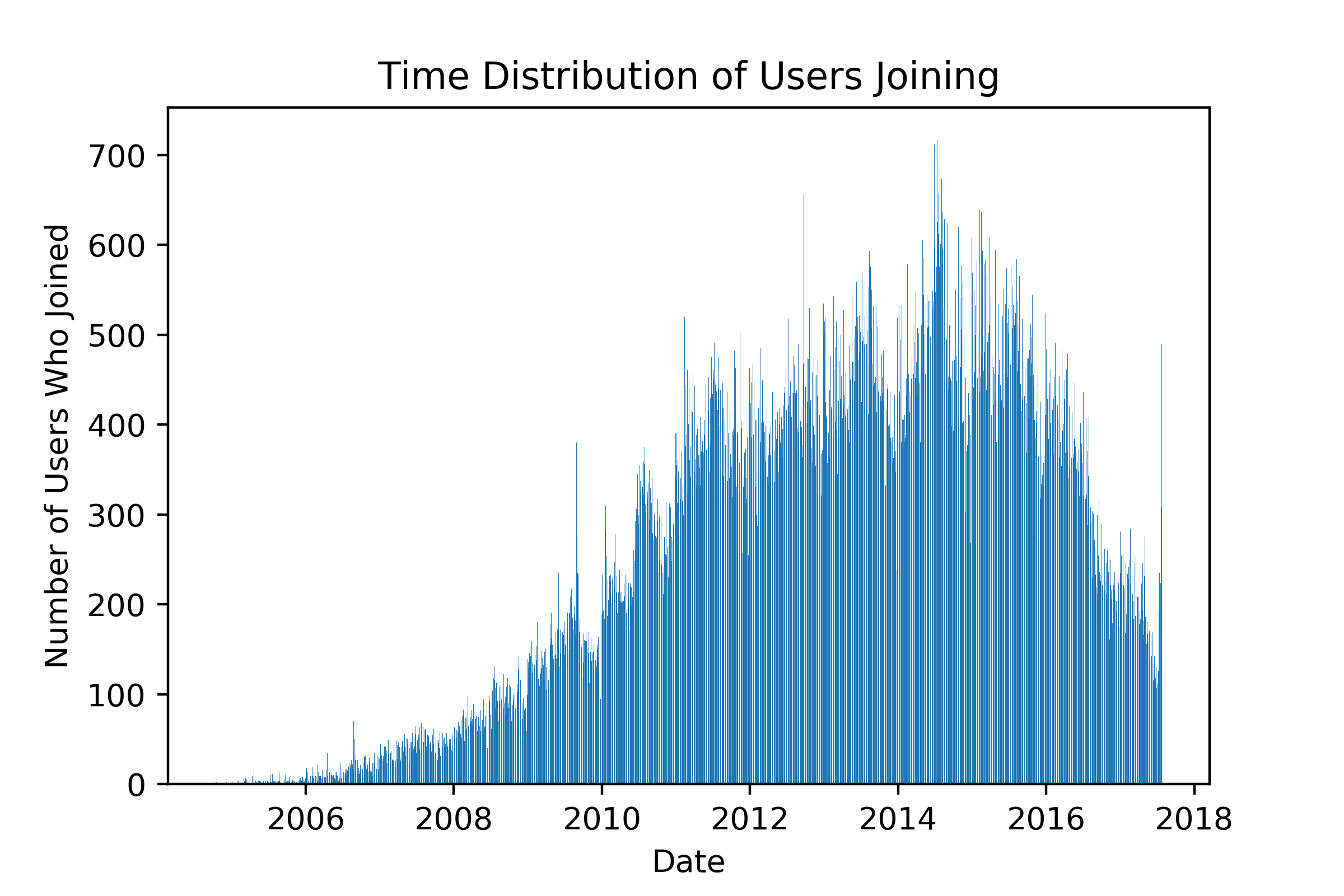}
\caption{Number of new users joining Yelp over time.}
\label{fig:users_join_over_time}
\end{figure}

\end{document}